\documentclass{article}
\usepackage[a4paper, total={6in, 10in}]{geometry}
\usepackage{authblk}
\usepackage{tabularx, booktabs}
\usepackage[hidelinks]{hyperref}
\usepackage{graphicx}
\usepackage{amsmath}
\usepackage{amssymb}
\usepackage{subfig}
\PassOptionsToPackage{hyphens}{url}\usepackage{hyperref}

\author[1]{Steven Stalder}
\author[1]{Michele Volpi}
\author[2]{Nicolas B\"uttner}
\author[3]{Stephen Law}
\author[2]{Kenneth Harttgen}
\author[4]{Esra Suel\thanks{Corresponding author.}}
\affil[1]{\normalsize Swiss Data Science Center, ETH Zurich and EPFL}
\affil[2]{\normalsize NADEL - Center for Development and Cooperation, ETH Zurich}
\affil[3]{\normalsize Department of Geography, University College London}
\affil[4]{\normalsize Centre for Advanced Spatial Analysis, University College London}
\date{}

\title{Self-supervised learning unveils change in urban housing from street-level images}

\begin{document}
\maketitle

\begin{abstract}
Cities around the world face a critical shortage of affordable and decent housing. Despite its critical importance for policy, our ability to effectively monitor and track progress in urban housing is limited. Deep learning-based computer vision methods applied to street-level images have been successful in the measurement of socioeconomic and environmental inequalities but did not fully utilize temporal images to track urban change as time-varying labels are often unavailable. We used self-supervised methods to measure change in London using 15 million street images taken between 2008 and 2021. Our novel adaptation of Barlow Twins, Street2Vec, embeds urban structure while being invariant to seasonal and daily changes without manual annotations. It outperformed generic embeddings, successfully identified point-level change in London’s housing supply from street-level images, and distinguished between major and minor change. This capability can provide timely information for urban planning and policy decisions toward more liveable, equitable, and sustainable cities.
\end{abstract}

\begin{figure}[t]
\centering
\includegraphics[width=\textwidth]{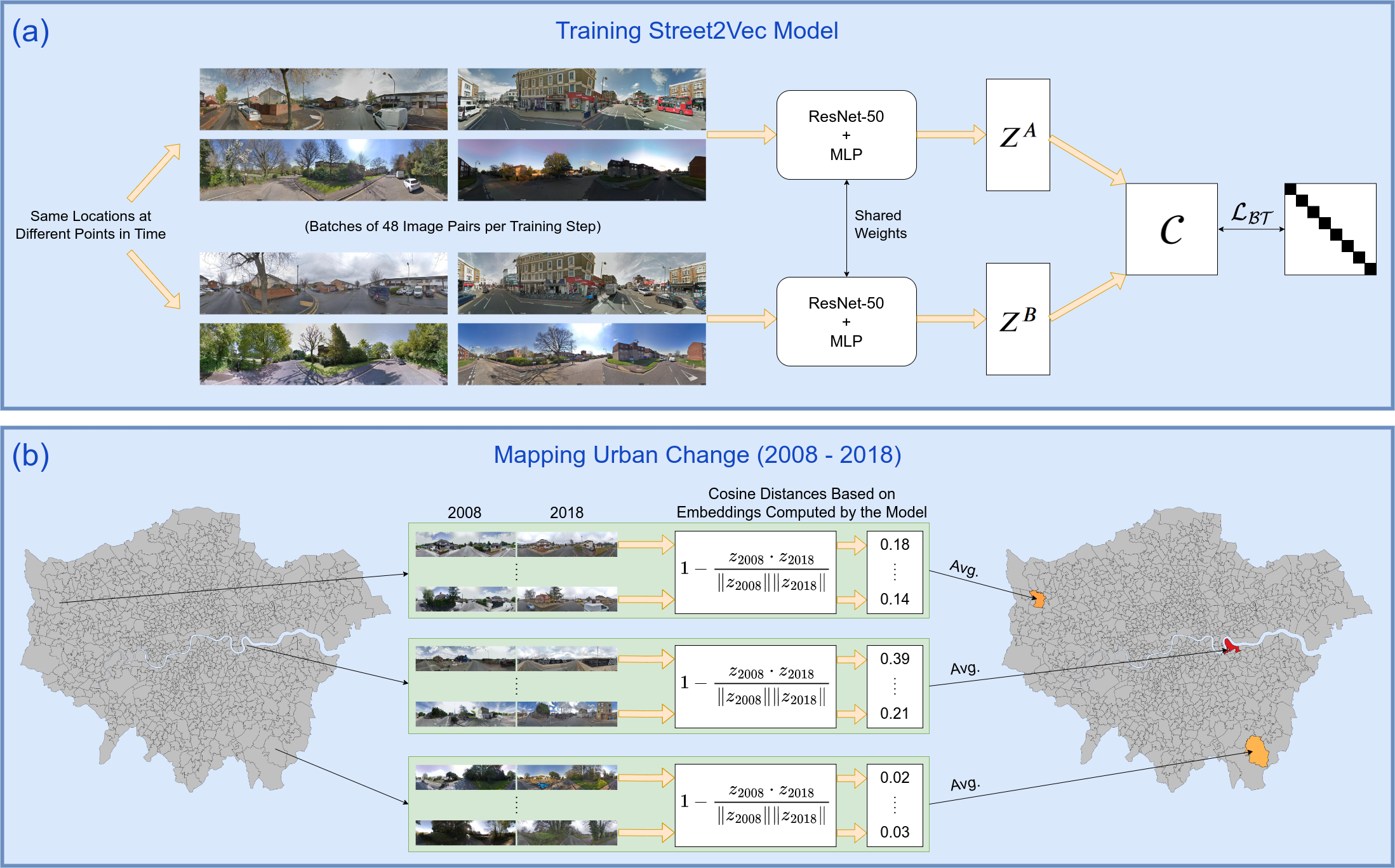}
\caption{Overview of the full pipeline. (a) Our proposed Street2Vec method, in which we apply Barlow Twins \cite{zbontar2021barlow} to street-level images. (b) Illustration of our application of the trained Street2Vec model for mapping urban change between 2008 and 2018.}
\label{Fig:method}
\end{figure}

The global urban housing crisis emerged as a pressing issue in recent decades, with cities worldwide facing a critical shortage of affordable and decent housing \cite{wetzstein2017global, saiz2023global, wef2022housing}. It has serious implications for health, social mobility, and economic productivity \cite{schuetz2022fixer}. Evidence suggests the lack of affordable high-quality housing exacerbates other inequalities, worsens homelessness, contributes to declining birth rates, and forces essential workers out of cities \cite{sachs2019six, quigley2004housing, shaw2004housing, yi2010effect, nyt2021essential, guardian2023sydney, guardian2021covidworkers, nyt2023newyork}. The United Nations (UN) recognizes the urgency; universal access to adequate, safe, and affordable housing is now part of the Sustainable Development Goals (SDGs) \cite{keith2023new}. City governments allocate resources to affordable housing initiatives and regenerating and expanding the housing supply. Timely measurements at high spatial resolution are crucial, yet largely lacking, for tracking progress and informing interventions \cite{brookings2022fixhousing}. Further research into methods allowing measurement of housing supply from emerging sources of low-cost large-scale data is essential to support timely data-driven decision-making for local governments and achieving global goals towards more liveable, equitable, and sustainable cities~\cite{fritz2019citizen}.

One of the barriers to effective tracking of housing is the fragmentation of housing data across disparate sources. The most comprehensive measure of housing stock is the census, yet it is only conducted every ten years in many countries. Although household surveys provide more frequent estimates of housing quality and costs, they lack spatial granularity due to their limited sample sizes. Local governments responsible for housing permits hold relevant data, yet building a unified dataset requires targeted collaborative efforts from multiple private and public sector players. In addition, new development data captures only a part of housing supply change and excludes demolitions, renewals, and regeneration. As a result, cities and researchers lack a comprehensive data source to effectively track the housing supply and its affordability to assess interventions as they are implemented. 

The availability of emerging sources of affordable large-scale image data, combined with advances in computer vision methods relying on deep learning, holds great potential for accelerating and improving urban measurements \cite{glaeser2018big}. Street-level images attracted particular attention as they capture urban environments as experienced by their residents and can provide very high spatial and temporal resolution~\cite{biljecki2021street, middel2019urban}. Prior research focused on the application of supervised deep learning methods to street-level images, which require high-quality ground truth measurements (labels) as a starting point. Successful urban applications included measurement of socio-demographics \cite{gebru2017using, suel2019measuring, suel2021multimodal}, trees and green space \cite{seiferling2017green, berland2017google, li2015assessing, long2017green, cai2018treepedia, Wegner_2016_CVPR}, housing prices \cite{law2019take}, crime rates \cite{suel2019measuring, mckee2017impact}, pollution levels and sources \cite{qi2021using, suel2022you, weichenthal2019picture, apte2017high}, perceptions \cite{naik2014streetscore, naik2017computer, larkin2022measuring, kruse2021places}, density and walkability \cite{garrido2023people, yin2015big, lu2018effect, doiron2022predicting, blevcic2018towards, nagata2020objective}, road safety \cite{mooney2020development}, accessibility \cite{hara2013combining}, and travel patterns \cite{goel2018estimating, zhang2019social, ibrahim2021urban, hankey2021predicting}. Even though mapping providers such as Google, Baidu, and Mapillary have been collecting and archiving multi-year street-level images of several cities for over a decade \cite{kim2023examination, anguelov2010google, goel2018estimating, biljecki2021street, neuhold2017mapillary, long2017green}, many of these studies were cross-sectional. So far, researchers have been limited by the difficulty of attaining temporally coherent and spatially dense label data at scale, as required by supervised methods. Therefore, existing research has not fully explored the potential of the temporal dimension of street-level images for studying urban change. 

On the other hand, self-supervised representation learning methods are being increasingly investigated as a way to extract meaningful information from large sets of structured but unlabelled data \cite{balestriero2023cookbook}. They also differ from traditional unsupervised methods (e.g., auto-encoders) \cite{law2019unsupervised} as they learn latent data representations by optimizing auxiliary tasks exposing intrinsic data structures, and have found greater success in discriminatory tasks \cite{wang2020urban2vec}. Previous works have applied these methods to aerial and satellite images \cite{jean2019tile2vec, ayush2021geography, manas2021seasonal, mall2023change} for land cover change detection \cite{volpi2015spectral, asokan2019change,dong2020self, saha2022self, chen2022self,wang2023self}. Comparatively little research applied self-supervised learning methods to street-level images where the focus has primarily been on learning image embeddings that contrast over geographical proximity and cross-modal embeddings for cross-sectional prediction tasks \cite{liu2023knowledge, wang2020urban2vec}. Measurement of change from street images holds value not only for data-rich cities that often lack timely measurement data but also for developing regions experiencing rapid urbanization and facing data scarcity~\cite{boeing2022using}.

In this study, we measured neighborhood change from street-level images using self-supervised representation learning (Fig. \ref{Fig:method}). We used 15.3 million images taken between the years 2008 and 2021 across the London metropolitan area in the UK. We applied our proposed Street2Vec method, where we adapted Barlow Twins \cite{zbontar2021barlow} to effectively learn from spatio-temporal street-level images. We designed Street2Vec to ensure it remains invariant to common but irrelevant variations across images such as illumination conditions, seasonality, and the presence of vehicles and people in images. We constrained its focus on learning visual features that relate to urban building structure. Once the Street2Vec representations were learned, we computed the degree of change based on the cosine distance between the embeddings of two images captured a decade apart, in 2008 and 2018, as this was the largest time span in our dataset for which we have a big overlap of image locations. We systematically assessed the performance of our Street2Vec approach. First, we visually and quantitatively evaluated change we detected in Opportunity Areas (OAs) that were announced and received financial incentives by the Mayor of London as key locations with potential for new homes, jobs, and infrastructure, and compared it to changes detected for all other areas in Greater London area. Second, we manually labeled 1,449 image pairs from 2008 and 2018 with respect to five different categories corresponding to degrees of urban change (see Supplementary Information for the detailed definitions). We then compared the change detection performance of our proposed Street2Vec approach with a baseline feature extractor pre-trained on ImageNet, which uses the same backbone Convolutional Neural Network (CNN) ResNet-50 architecture \cite{he2016deep}. Finally, we visualized the spatial distribution of neighborhood clusters from Street2Vec embeddings.

Our models successfully identified both subtle and significant urban transformations corresponding to changes in housing stock at the neighborhood level in London, outperforming the generic pre-trained feature extractor. Our change detection results are based on 329,031 point locations in London for which we had images from both 2008 and 2018.

\section*{Results}

\begin{figure}

\begin{minipage}{.5\linewidth}
\centering
\subfloat[]{\label{main:a}\includegraphics[scale=.14]{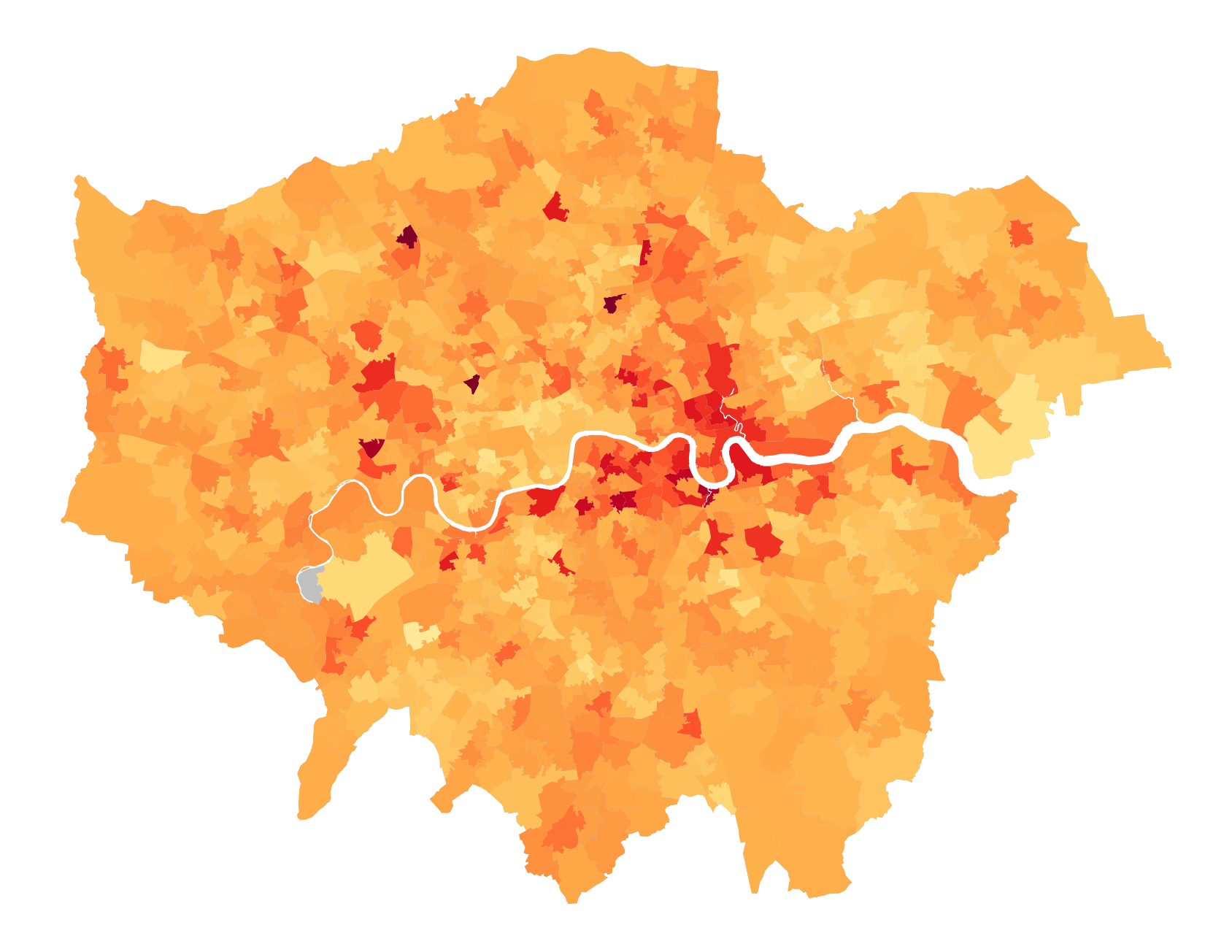}}
\end{minipage}%
\begin{minipage}{.5\linewidth}
\centering
\subfloat[]{\label{main:b}\includegraphics[scale=.14]{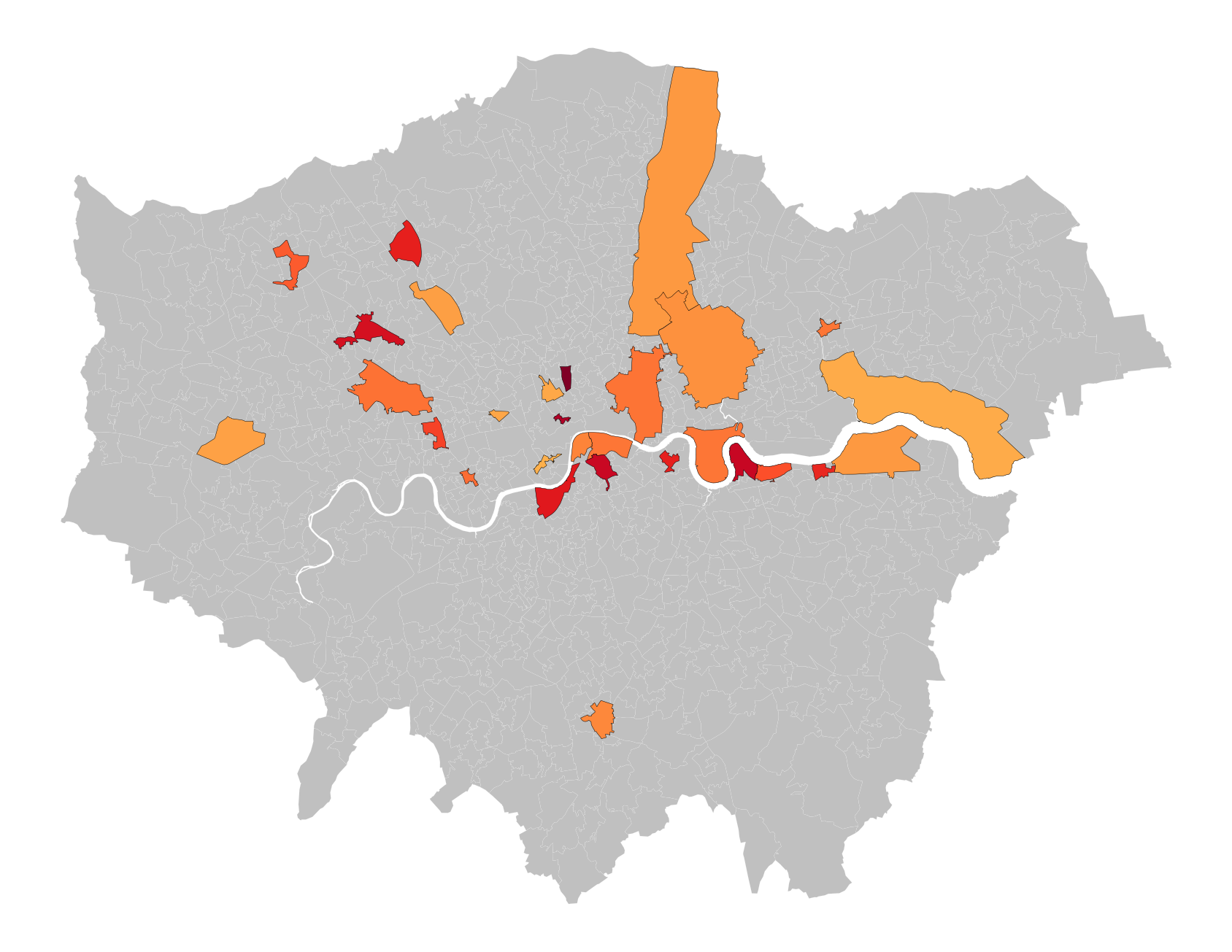}}
\end{minipage}\par\medskip
\centering
\subfloat[]{\label{main:c}\includegraphics[scale=.5]{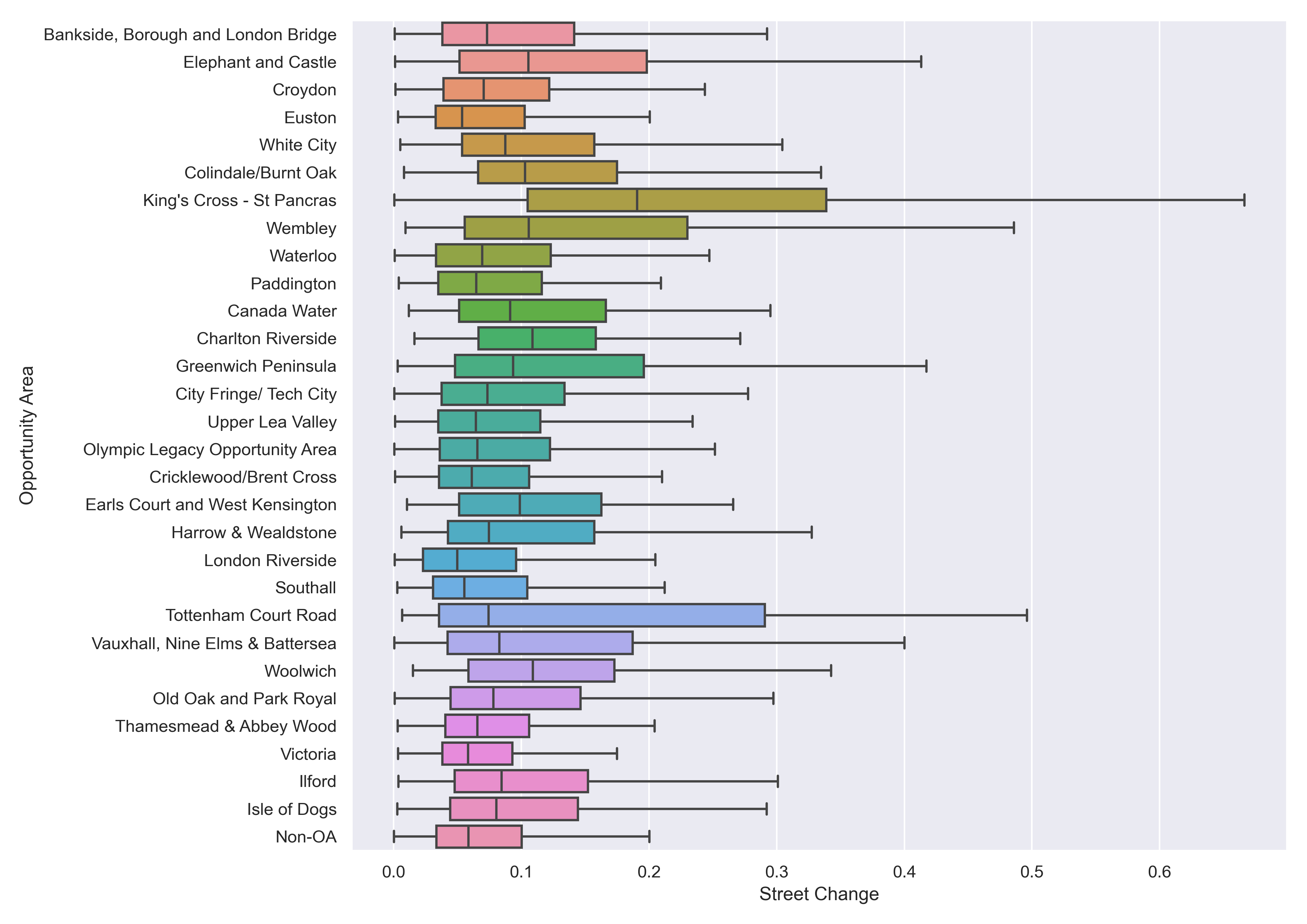}}

\caption{Change detected from Street2Vec embeddings in London: (a) map of predicted mean change for all Middle Super Output Areas (MSOAs), (b) map of predicted mean change for Opportunity Areas (OAs) announced by the Mayor of London as areas with substantial potential for new developments, (c) distribution of point level change detected in OAs in London compared with non-opportunity areas (Non-OA) in London. In (a) and (b), darker red colors correspond to higher levels of predicted change.}
\label{Fig:main}
\end{figure}

\subsubsection*{Change in Opportunity Areas}
As part of the spatial development strategy published and periodically updated by the Mayor of London since 2004, Opportunity Areas (OAs) in Greater London are identified as key locations with potential for new homes, jobs, and infrastructure (\url{london.gov.uk/programmes-strategies/planning/implementing-london-plan/londons-opportunity-areas}). New developments in these neighborhoods have been actively incentivized by the Mayor and Local Authorities through various measures including investments in transport links. To evaluate the effectiveness of our model, we estimated the point-level change between 2008 and 2018 as measured by cosine distances between Street2Vec embeddings. We expected to see higher numbers of locations with larger change within OAs (i.e., higher values for the median and 75th quantile). We show a comparison of distributions of point-level change in Fig.~\ref{Fig:main} for all OAs separately along with all other areas combined. We found that OAs have significantly higher levels of median change ($p<0.01$). We also found that OAs have significantly higher values for the 75th quantile when compared to other areas in London ($p<0.01$). While we do not have ground truth data to validate point-level change detection, our results demonstrate the success of our model in highlighting neighborhoods where we would expect the largest change in London. Our findings revealed substantial variation as captured by cosine distances, with certain areas experiencing substantial neighborhood change (e.g., Kings Cross and St. Pancras, Tottenham Court Road), along with areas lying on newly built transportation infrastructure such as the Northern Line extension or the Elizabeth Line (e.g., Battersea and Woolwich). Other areas showed limited change or are lagging in planning. This also provides important information for the local governments as it highlights areas that have not experienced the anticipated level of development despite incentives enabling targeted interventions, and also ones that have organically redeveloped without targeted policy interventions.

\subsubsection*{Subtle shifts vs. major new developments}
We investigated whether our method can also identify relatively subtle changes in the built environment, for instance from the renewal of existing homes to regeneration of entire neighborhoods. We expected major new housing to have stronger visual signals from images. However, a lot of the change in housing happens through shifts in existing housing stock, especially in cities with rich histories like London. This type of change may show weaker visual signals in street-level images, potentially making it more challenging to monitor effectively. Tracking subtler changes, such as new coffee chains or repainting of facades, is critical as they may lead to undesired outcomes such as displacement caused by processes like gentrification.

\begin{figure}[t]
    \centering
    \includegraphics[width=\textwidth]{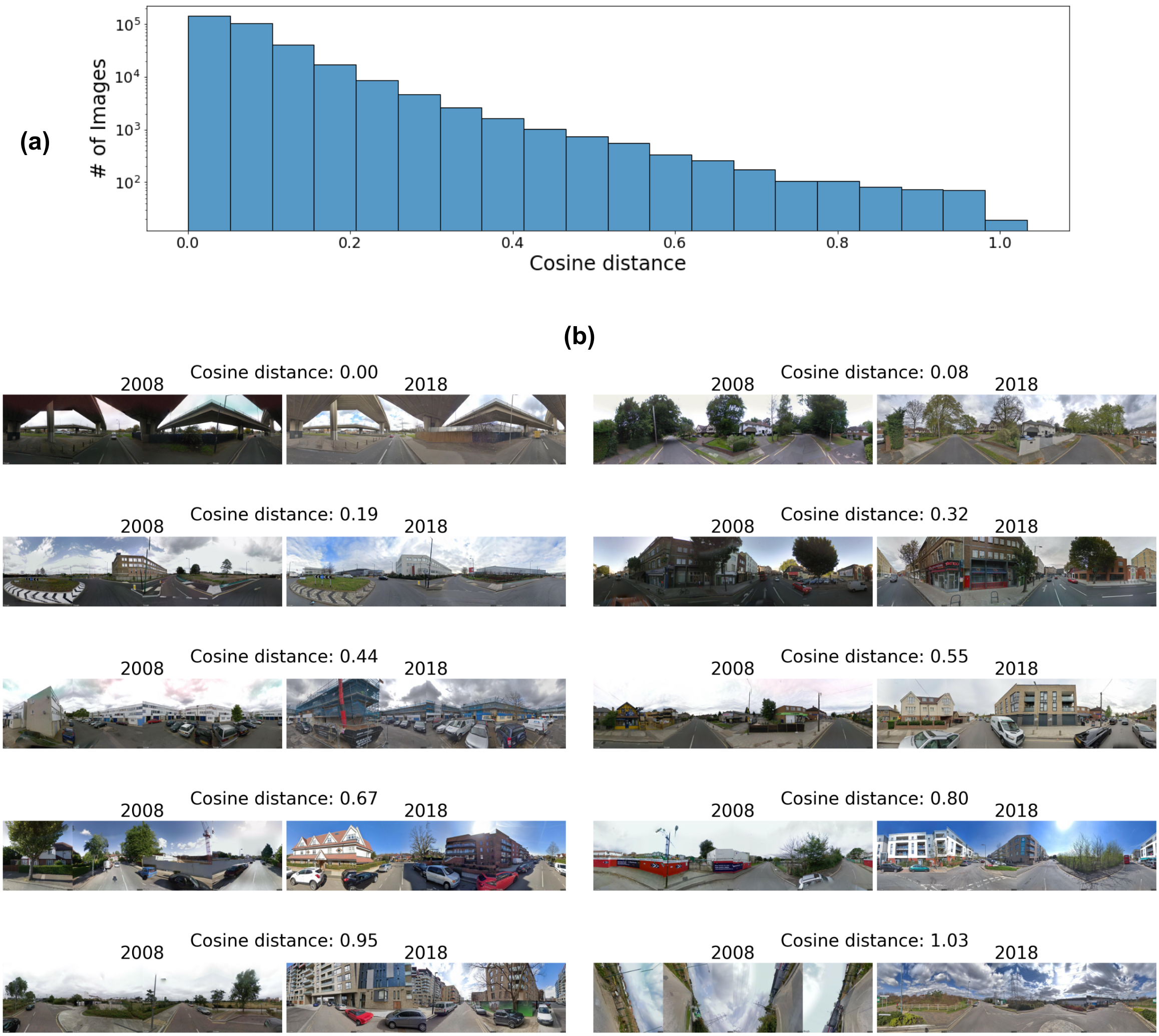}
    \caption{(a) Histogram of cosine distances between the embeddings of images in 2008 and 2018. (b) Example image pairs with increasing cosine distances between them.}
    \label{Fig:cos_dist_examples}
\end{figure}

A visual investigation of image examples such as in Figure~\ref{Fig:cos_dist_examples}(b) revealed that our model can indeed distinguish between different levels of structural change in the built environment. For example, the first image pair demonstrates that our model attributes close to zero change to pairs that show considerable visual differences in lighting conditions or seasonality. Minor structural alterations like the construction of fences such as in the second image pair of the first row lead to small but non-zero cosine distances. The second row shows examples of image pairs with slightly higher cosine distances, and they indeed begin to exhibit stronger changes such as a new paint job on the facade and the widening of the pavement. Newly constructed buildings are visible in the third row, while the fourth row with very high cosine distances captures complete reconstructions of streets or neighborhoods. The fifth row shows examples of pairs with the highest distance values, where we found instances of extreme structural change and also many images that were rotated, corrupted, or anomalous in other ways. All of this was desired behavior when designing our proposed Street2Vec method.

As we did not have suitable ground truth data for a quantitative evaluation, we manually labeled 1,449 image pairs from the years 2008 and 2018 according to stratified random sampling based on five value ranges from our model predictions. We used five classes for labeling, representing an ordinal scale of change between street-level panoramas: (1) minimal irrelevant change, (2) noticeable but irrelevant change, (3) minor urban change, (4) major urban change, and (5) anomalies in images (see Supplementary Information for detailed explanations). After labeling the image pairs according to our classification, we computed the empirical histogram of the cosine distances for each class. To have a basis for comparison, we also extracted representations and computed corresponding cosine distances from a baseline CNN architecture \cite{he2016deep}, which was only trained for classification on the ImageNet dataset~\cite{deng2009imagenet} and has not been trained to learn urban representations. This test aims to assess whether Street2Vec has indeed learned more suitable representations for capturing urban change, or if a simpler yet coherent visual representation would have sufficed to distinguish between varying levels of change.

\begin{figure}[t]
    \centering
    \includegraphics[width=\textwidth]{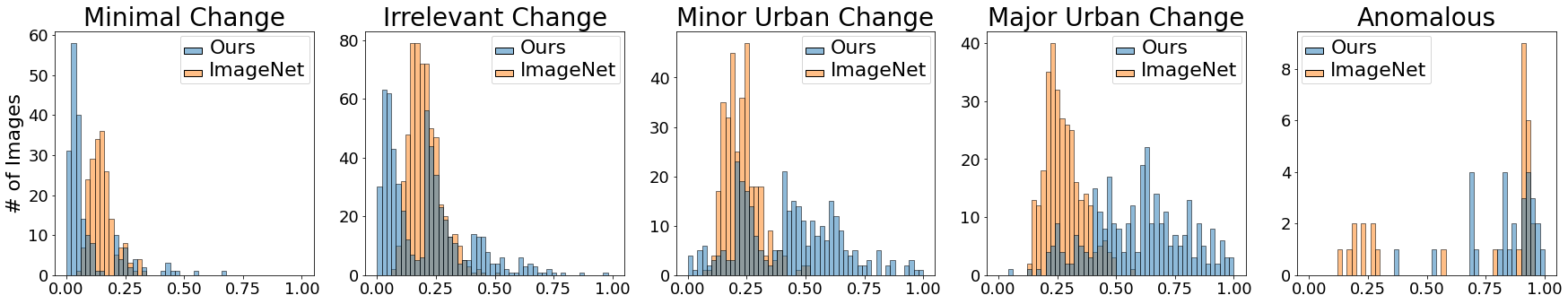}
    \caption{Histogram of cosine distances per class, for our model (blue) and for the baseline (orange).}
    \label{Fig:label_hist}
\end{figure}

We found that the mean cosine distances from Street2Vec followed the expected order, successfully distinguishing between minimal, irrelevant, minor, and major change (Table 1). While there is overlap between classes (Fig.~\ref{Fig:label_hist}), these results demonstrate that our model learns visual features of relevant urban change, corresponding to semantics related to urban structures while disregarding (or putting less emphasis on) irrelevant visual differences. Street2Vec embeddings also outperformed generic features learned from the baseline CNN model. The mean distances from the baseline CNN model also followed the correct order, however, the cosine distance distribution did not show the desired spread which would allow to accurately separate change classes. One notable observation regarding Street2Vec embedding distributions is the emergence of two peaks for the ``irrelevant change'' and ``minor urban change'' classes in Fig.~\ref{Fig:label_hist}. While we cannot determine with certainty why these occur, this may be related to intra-label variances and our own biases during the labeling process. Finally, as expected, both models effectively identify anomalous images.

\begin{table}[t]
\caption{Comparison of mean cosine distances per class for embeddings from Street2Vec and the baseline CNN model pre-trained on ImageNet.}
\label{Tab:labeling_results}
\centering
    \begin{tabular}{@{}lrrrrr@{}}
    \toprule
     & Minimal  & Irrelevant  & Minor urban  & Major urban  & Anomalous\\
     & Class 1 & Class 2 & Class 3 & Class 4 & Class 5\\
    \midrule
    Street2Vec & 0.090 & 0.205 & 0.424 & 0.592 & 0.838\\
    Baseline & 0.151 & 0.201 & 0.228 & 0.275 & 0.685\\
    \bottomrule
    \end{tabular}
\end{table}

\subsubsection*{Visualizing the neighbourhood clusters}

\begin{figure}[t]
\centering
\includegraphics[width=0.95\textwidth]{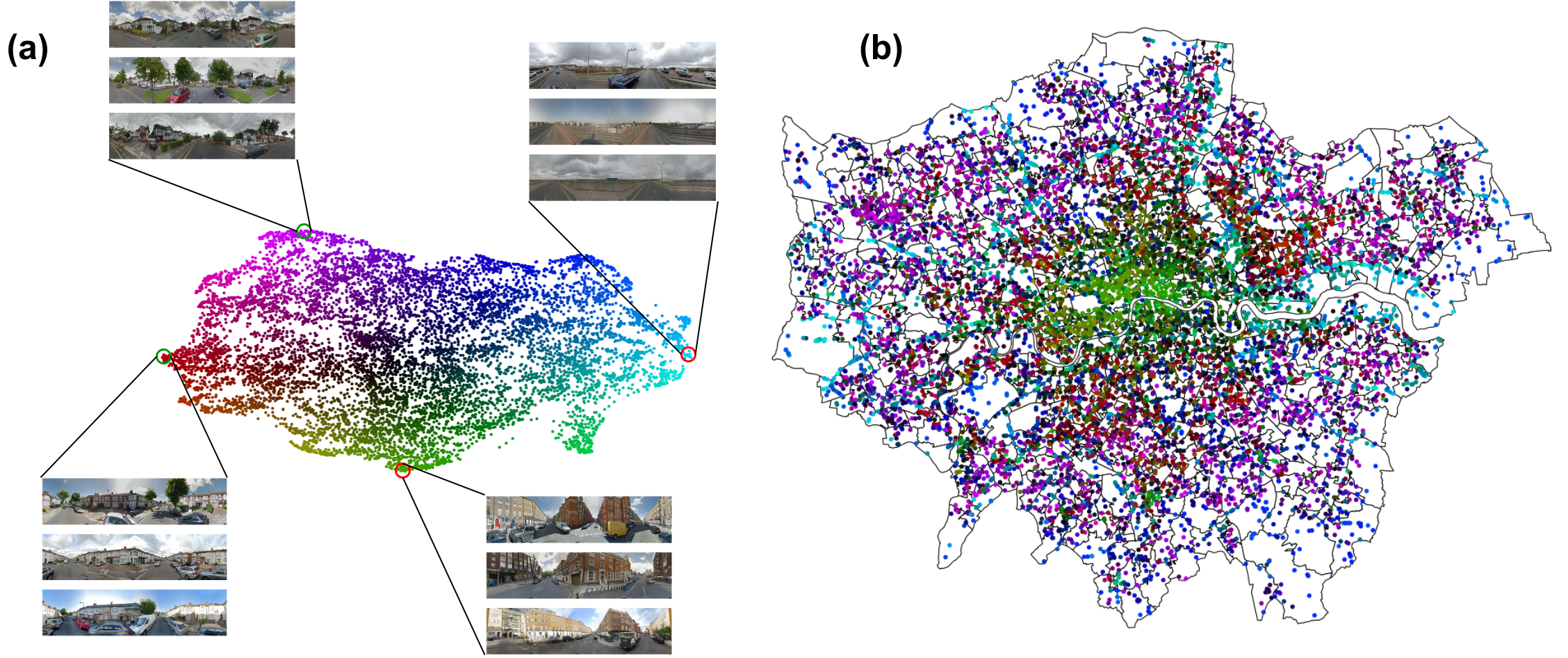}
\caption{(a) UMAP~\cite{mcinnes2018umap-software} projection space for 10,000 randomly sampled street-level images from London that have been processed by Street2Vec. We color the points in a circular manner according to their position in this space. Note that we omit axes and scales, as the absolute values of the UMAP embedding are meaningless and the plotting is purely qualitative. On the extremes of the two UMAP dimensions, we plot the three images corresponding to the minimal or maximal values, respectively. (b) The same points on the map of London. Similar colors mean that two data points are close according to the learned representations.}
\label{Fig:umap_circ_colors}
\end{figure}

To further interpret the learned embeddings from Street2Vec, we visualized their spatial distribution using 10,000 randomly sampled street-level panoramas (taken at any point in time to get unbiased temporal and spatial coverage) based on their position in a Uniform Manifold Approximation and Projection (UMAP) \cite{mcinnes2018umap-software} to two dimensions (see \nameref{sec:methods}). Similar colors in Fig.~\ref{Fig:umap_circ_colors} can be interpreted to represent similar neighborhoods according to our model.

Our coloring reveals interesting spatial patterns even though geographical coordinates were never explicitly given as a model input. In Fig.~\ref{Fig:umap_circ_colors}(b), the city center is clustered together in different shades of green. Light green colors gradually change to dark green, to red, to blue and magenta colors moving outward from the center towards the suburbs. The light blue points appear to follow London motorways. To provide an intuitive understanding of the information captured by UMAP's two embedding dimensions, Fig.~\ref{Fig:umap_circ_colors}(a) showcases sample images situated at the farthest extremes of the UMAP latent dimensions distribution. In the first (horizontal, x-axis) dimension, the three images with the lowest values come from residential areas with low-rise buildings (red points in the map), and the three images with the highest values are from roads and do not show any nearby settlements (light blue points in the map). Therefore, a possible interpretation for the first dimension of the UMAP could be ``habitableness''. In the second (vertical, y-axis) dimension, images with the lowest values come from the urban core near the center (light green points in the map), while the ones with the highest values are suburban-looking scenes with noticeable vegetation content (magenta points in the map). It seems likely that this dimension captures some form of ``urbanization'' in the images. However, such visual interpretations are qualitative and UMAP projections from a high-dimensional space down to two dimensions are difficult to interpret as they are bound to lose information, whereas the overall modes of variations in the street-level images learned by Street2Vec are likely much more complex. 

\section*{Discussion}
To our knowledge, this is the first application of self-supervised deep learning methods to demonstrate successful use of temporal street-level images for measuring urban change. We have shown that street-level images can capture changes in the built environment without the need for manual labels. We also found that our models can distinguish between changes in housing from regeneration and renewal of existing homes that may have weaker visual signals (minor urban change), and larger housing development projects (major urban change). Our approach can be readily applied to existing street-level image datasets, which are already available worldwide and undergo periodic updates by commercial providers. While researchers currently face access restrictions, the demonstrated success of our proposed method has the potential to capture the attention of data owners. This could lead to cost-effective partnerships or integration into existing data pipelines, resulting in the creation of a comprehensive global dataset. Consequently, an invaluable tracking tool would be developed, facilitating the measurement of progress toward achieving universal access to adequate, safe, and affordable housing on a global scale. This tool would not only benefit local governments and countries but also contribute significantly to the attainment of SDGs.

The study has several strengths. It used a publicly available large dataset of street-level images being consistently collected since 2007. This image dataset had substantial spatio-temporal coverage on an urban scale for the Greater London metropolitan area with sufficient power to investigate neighbourhood-level progress. The study employed our proposed Street2Vec adapted from the Barlow Twins \cite{zbontar2021barlow}. This self-supervised learning technique has demonstrated its efficacy in pre-training expansive models for multiple computer vision tasks such as image classification, object detection, and image segmentation. Prior research on image-based urban measurement methods relied on the availability of high-quality labels, which are often lacking. In fact, our motivation for developing an image-based proxy to characterize urban change stemmed from the scarcity of ground truth data, even in data-rich cities of the developed world. The main novelty of our proposed measure of urban change is that it does not require labels, and only makes use of street-level images and its readily available metadata on acquisition year and location. Our study demonstrated the effectiveness of the proposed Street2Vec approach by quantitatively comparing the detected change in London neighborhoods designated by the city for potential new housing and job opportunities, with others. For example, we found that the housing policy and investments have been successful in initiating change in many OAs in London, such as King's Cross, while not all OAs experienced similar outcomes. This analytical capability is essential for informing local governments, particularly for tracking areas where existing incentives have fallen short in stimulating housing development, as well as for identifying organic changes that extend beyond policy-driven areas and are therefore challenging to monitor. Furthermore, our approach successfully identifies subtle urban change from the regeneration and the renewal of existing homes and neighborhoods. These changes may exhibit weaker visual signals, yet are crucial for cities like London with rich histories and aging housing stocks. In such cities, regeneration efforts may inadvertently lead to undesirable outcomes such as population displacement caused by processes of gentrification.

A number of potential limitations could have influenced our results and could impact a wider adoption of our proposed method as a standalone tracking tool. First, the assumption that all changes in the housing stock will have visual signals captured by street images may not always hold because renewal projects such as energy efficiency improvements or increasing housing capacity within existing buildings, or differences in uses may not always result in visible changes in external views. Therefore, ideally, image-based tracking should be combined with other data sources to address such weaknesses. In addition, anomalous images as well as rare weather events such as snow in London influence change detection results that need to be taken into account for interpretations. Thereby, future research can focus on improving the discriminatory power of Street2Vec for these rare occurrences, but also improving the detection of specific urban change, using potentially weak labels (with minimal or cheap human intervention) or generative models to improve self-supervised learning. Street-level imagery is predominantly collected and controlled by major private sector players (e.g., Google, Baidu, Bing) providing extensive coverage globally. However, they are also increasingly imposing access restrictions, even for non-commercial uses intended for the public good. Cities can require or incentivize better access, support crowd-sourced initiatives, or use their own assets to enhance monitoring. The coverage of collected data may be biased towards areas with substantial urban change as the original purpose of street-level images was to provide up-to-date mapping services, but this is less problematic as the it is well aligned with our primary interest on change. Still, coverage may be problematic in some countries where there is high public scrutiny of data privacy (e.g., Germany) and restricted access in developing areas such as slums. While our approach is generalizable to most cities around the world, accessing data for evaluating local government performances may present greater challenges, as London's geospatial data initiatives are considered among the most comprehensive and advanced in the world. Finally, further research is needed to better understand how learned representations from Street2Vec could be used for other downstream tasks of critical importance for city governments such as tracking climate resilience, affordability, and displacement. 

\section*{Methods}\label{sec:methods}
\subsubsection*{Image data}
We used a total of 15,335,000 images at 3,833,750 locations in London spanning the years from 2008 to 2021. We accessed the images using the Google Street View Application Programming Interface (API). We first created a 50m grid using the street network obtained from Open Street Maps (OSM) within the city boundary shape file for the Greater London Authority. We then used the API to retrieve the unique panorama ids (panoIDs) of images near each grid point acquired by Google. For each sampled panoID location, we used four images (of size 600x600 pixels) representing four orientations of a 360$^\circ$ panorama (0$^\circ$, 90$^\circ$, 180$^\circ$, 270$^\circ$) to cover all directions within a view. For change detection, we used images from all 329,031 locations for which we had images from both 2008 and 2018.

\subsubsection*{Street2Vec: Self-supervised learning from street-level images}
In our setting, we have a large set of geolocated images, for which the only information available is the acquisition year and location. When labels are available, it is common to directly train a fully supervised model approximating the function of interest. However, in the present study, we are interested in estimating the degree of urban change between two images from the same location taken at different years where we don't have ground truth (label) data. This problem setting is particularly well suited for a relatively recent paradigm in machine and deep learning, known as \emph{self-supervised representation learning} (SSL in short). The goal of SSL is to learn vector representations (i.e., \emph{embeddings}) of inputs that can represent the relevant information in a condensed form, without the need for target labels. The learning problem is set up in a way that a model can be learned using standard supervised losses (e.g., cross-entropy, mean squared error, etc.) using surrogate labels generated from the data itself rather than from an external source (hence, self-supervised learning). Our goal is to learn representations of street-level images that can capture and discover structural urban change but are invariant to common but irrelevant variations such as lighting conditions, seasonality, or vehicles and people appearing in the images. To achieve this, we propose an adaptation of the Barlow Twins method from \cite{zbontar2021barlow} to temporal street-level images: Street2Vec.

In the original paper~\cite{zbontar2021barlow}, the Barlow Twins method is applied to two different sets of content-preserving artificial distortions on images, such as cropping, color jittering, and horizontal flipping, among others. The model is trained by forcing it to learn the same representations for these different sets of modified images. This is semantically coherent since every batch of images is transformed to obtain two slightly different versions of the same images that do not alter the relevant content in them. Both distorted batches are fed through a ResNet-50 model \cite{he2016deep}, and a Multi-Layer Perceptron (MLP, also known as fully connected layers) projector network, consisting of three layers with batch normalization and rectified linear unit (ReLU) nonlinearities between them. The resulting two batches of embedding vectors $Z^A$ and $Z^B$ are first standardized to zero mean and unit standard deviation along their batch dimensions, and then multiplied among them to compute the cross-correlation matrix $\mathcal{C}$ of the embedding feature dimensions, averaged over the batch dimension: 

\begin{equation}
\label{eq:bt_loss}
    \mathcal{L}_\mathcal{BT} \triangleq \sum_i^D (1 - \mathcal{C}_{ii})^2 + \lambda \sum_i^D \sum_{j \neq i}^D\mathcal{C}_{ij}^2 \; \text{,}
\end{equation}

\noindent where $D$ is the embedding feature dimension, $\mathcal{C}$ is the cross-correlation matrix, and $\lambda$ is a tunable hyperparameter controlling how much off-diagonal correlations are penalized. The cross-correlation matrix $\mathcal{C}$ is defined as follows:

\begin{equation}
\label{eq:cross_corr}
    \mathcal{C}_{ij} \triangleq \frac{\sum_b^N z_{b,i}^A z_{b,j}^B}{\sqrt{\sum_b^N (z_{b,i}^A)\strut^2}\sqrt{\sum_b^N (z_{b,j}^B)\strut^2}} \; \text{,}
\end{equation}

\noindent where $N$ is the batch dimension, and $z_{b,i}^A$ and $z_{b,j}^B$ are elements of the mean-centered embeddings $Z^A$ and $Z^B$.

The loss function in Eq.~\eqref{eq:bt_loss} serves the purpose of learning image embeddings where the different dimensions are uncorrelated (i.e., each representing some ``new'' properties of the data that are unrelated to the others) but the entries of the same embedding dimension from the two batches are maximally correlated. The intuition behind this second part is that we want the embeddings to be invariant to the specific distortions applied to the input image. Through this method, the model is learning to represent relevant image contents in vectorized form.

In our setting, instead of applying a set of predefined artificial distortions to the images, we take two images from the same location, captured in different years as input. That is, at each training step, we sample a new batch of street-level images from random locations and points in time and then sample the second batch of images (the ``distorted'' samples) from the same locations, but taken in a different year if at least two images from different years are available for the selected coordinates. In the rare case where we only have one image from a single year at a given coordinate location, we simply apply some small amounts of color jittering (i.e., small random changes in brightness, contrast, saturation, and hue) to the sampled image, to obtain a second, artificially distorted one. After that, we apply the Barlow Twins method as explained above, to learn aligned image embeddings with uncorrelated feature dimensions. See Fig. \ref{Fig:method} for an overview. We name this approach Street2Vec, as we learn visual vector embeddings from street-level images.

Our assumption is that on average, street-level images taken at two different time instants will have strong visual appearance variations representing changes that are not the focus of this study such as lighting conditions, seasonality, people, or cars, but no or only minimal change in urban structural elements. Of course, we cannot completely rule out any structural change between any two of those images and in fact, our primary interest is to identify locations where structural change is captured by street-level images. However, we expect that these cases are occur much less frequently. Therefore, we posit that our model implicitly learns representations that are invariant to irrelevant change, but sensitive to urban structural elements, without labels explicitly highlighting such changes. We define irrelevant change to include lighting conditions, seasonal change in vegetation or clouds, snow, change in the view resulting from the relative position of the camera, and occlusion of built environment features by cars, vegetation, or individuals (see Supplementary Information for more detailed descriptions).

We trained our model on a single GPU, performing one pass over all our data, which took about 30 hours to complete. Longer training has not resulted in noticeably improved model performance. To maximize the information content within each model input, we concatenate all four available street-level image orientations (north, east, south, west) into a single panoramic view. Each orientation of size 600x600 pixels is resized to 128x128 pixels, resulting in a size of 128x512 pixels when we concatenate all four orientations that we give as one input to our model. Because of memory limitations, the largest batch size we could use was 48. Moreover, we used an embedding dimensionality of 1024 and kept $\lambda = 0.005$, as in \cite{zbontar2021barlow}.

Once the model is learned and the training converges, we then project every image into the learned embedding space. To perform change detection, we compare single image embeddings at all locations of interest and extract a single summary statistic representing a notion of deviation or distance. The farther the embeddings are, the more likely the location represented by the pair of images is to have undergone structural urban changes. Conversely, the closer the embeddings are, the more likely the images are to either not have changed or to display only irrelevant change which the model has learned to become invariant to. 

\subsubsection*{Measurement of change}
Our main objective is to utilize the learned embeddings to detect structural changes in London neighborhoods captured by street-level imagery. To analyze our ability to perform this task, we selected two years, 2008 and 2018, that are furthest spaced apart among the years for which we had a considerable amount of image pairs from the same locations available. For every location where we have images for 2008 and 2018, we compute the cosine distance ($d_{\text{cos}}(\cdot, \cdot))$ between their Street2Vec embeddings as our change metric. The cosine distance between two vectors $\mathbf{x}$ and $\mathbf{y}$ is defined as 1 minus the cosine similarity ($s_{\text{cos}}(\cdot,\cdot)$), as follows:

\begin{equation}
    d_{\text{cos}}(\mathbf{x}, \mathbf{y}) = 1 - s_{\text{cos}}(\mathbf{x}, \mathbf{y}) = 1 - \frac{\mathbf{x}^T \mathbf{y}}{\lVert \mathbf{x} \rVert_2 \lVert \mathbf{y} \rVert_2} \; \text{,}
\end{equation}

\noindent and ranges from 0 (the vectors $\mathbf{x}$ and $\mathbf{y}$ are perfectly collinear and codirectional, the angle between them is 0) to 2 ($\mathbf{x}$ and $\mathbf{y}$ are completely opposite and the angle between them is $\pi$). However, note that in our setting, the maximum cosine distances are around 1 ($\mathbf{x}$ and $\mathbf{y}$ are orthogonal). We used the cosine distance because we have relatively high-dimensional embeddings where other distance metrics like Euclidean distances would be very sensitive to large deviations in only a few dimensions, even if the two embeddings would be very similar in most other dimensions. Since we assume each of our uncorrelated embedding dimensions to capture (equally) important information, we prefer capturing some change in many of them to capturing a lot of change in only a few.

\subsubsection*{Clustering neighborhoods}
We used a non-linear dimension reduction technique like UMAP instead of a linear dimension reduction technique like Principal Component Analysis (PCA) \cite{pearson1901pca}. In the Street2Vec representation learning method, we minimize a learning objective that, if perfectly optimized, decorrelates all feature dimensions of the embeddings. In such case, PCA would not be able to find a projection of the data that summarizes any meaningful information (in terms of variance) and it would reduce to a simple rotation of the embeddings. Even though perfect decorrelation will never be achieved in practice, it is still substantial enough that the eigenvalues of the first two principal components are only marginally larger than the reciprocal of the embedding dimensionality. For this reason, we employ the nonlinear manifold learning technique UMAP, which is able to find more complex relationships on the manifold of the learned representations of our geolocated images while meaningfully projecting into a lower-dimensional space (2-dimensional in this case). In that space, two neighboring points have similar embeddings, while two far apart points likely have strongly different ones. Therefore, for each street-level planar panorama, two close-by data points represent similar urban structures, according to our model.

\section*{Data and code availability}
Datasets used in this paper are publicly available and sources are provided in the main manuscript. Manual labeling The code for Street2Vec training and creating the change metric is available at \url{https://gitlab.renkulab.io/deeplnafrica/Street2Vec}

\section*{Competing interests}
The authors declare no competing interests.

\bibliography{street2vec}

\newpage
\section*{Supplementary Information}

\subsection*{Detail of manual annotations}\label{sec:app_classes}

In the following bullet points, we detail the criteria used to classify 1,449 image pairs (from 2008 and 2018) into one of 5 classes. 

\begin{itemize}
    \item Class 1: Minimal irrelevant change
    \begin{itemize}
        \item No urban change, only minimal changes in lighting/seasonality/relative position of the images
        \item No occlusion of buildings through traffic, vegetation, etc.
    \end{itemize}
    \item Class 2: Noticeable, but irrelevant change
    \begin{itemize}
        \item Snow
        \item Clear changes in lighting, seasonality or other irrelevant changes not relating to actual urban change (e.g., different sign from the same store, or one bike shop replacing another, etc.)
        \item (Partial) occlusion of buildings through traffic, vegetation, etc.
        \item Non-minimal change in position that the images have been taken in (sometimes this effect could occlude unchanged structures)
        \item Minor changes to landscaping
    \end{itemize}
    \item Class 3: Minor urban change
    \begin{itemize}
        \item (Relevant) storefront changes (e.g., store replaced by fancy restaurant)
        \item Building renovations
        \item New building in the same architectural style (e.g., new house in suburbs that looks similar to the others)
        \item Major changes to landscaping not related to seasonality, significant changes in urban furniture
    \end{itemize}
    \item Class 4: Major urban change
    \begin{itemize}
        \item Many new buildings for the entirety of the scene
        \item New building in different architectural style and typologies (e.g., larger house/skyscraper appearing, store replacing residential buildings, etc.)
    \end{itemize}
    \item Class 5: Anomalous
    \begin{itemize}
        \item No image
        \item Rotated images
        \item (Almost) completely dark images
        \item Other anomalies
    \end{itemize}
\end{itemize}
\end{document}